RODOLFO DELMONTE

# Expressivity in TTS from Semantics and Pragmatics

In this paper we present ongoing work to produce an expressive TTS reader that can be used both in text and dialogue applications. The system called SPARSAR has been used to read (English) poetry so far but it can now be applied to any text. The text is fully analyzed both at phonetic and phonological level, and at syntactic and semantic level. In addition, the system has access to a restricted list of typical pragmatically marked phrases and expressions that are used to convey specific discourse function and speech acts and need specialized intonational contours. The text is transformed into a poem-like structures, where each line corresponds to a Breath Group, semantically and syntactically consistent. Stanzas correspond to paragraph boundaries. Analogical parameters are related to ToBI theoretical indices but their number is doubled. In this paper, we concentrate on short stories and fables.

## 1. *Introduction*

In this paper we present ongoing work to produce an expressive TTS reader that can be used both in text and dialogue applications. The system called SPARSAR has been used to read (English) poetry so far but it can now be applied to any text provided it has been previously fully analyzed both at phonetic and phonological level, and at syntactic and semantic level. In addition, the system has access to a restricted list of typical pragmatically marked phrases and expressions that are used to convey specific discourse function and speech acts and need specialized intonational contours. In this paper, we concentrate on short stories and fables.

Current TTS systems are dull and boring and characterized by a total lack of expressivity. They only take into account information coming from punctuation and in some cases, from tagging and syntactic constituency. Few expressive synthetic speech synthesizers are tuned to specific domains and are unable to generalize. They usually convey specific emotional content linked to a list of phrases or short utterances – see below. In particular, comma is a highly ambiguous punctuation mark with a whole set of different functions which are associated with specific intonational contours. In general, question and exclamative marks are used to modify the prosody of the previous word. We use the word "expressivity" in a specific general manner which includes sensible and sensitive reading that can only be achieved once a complete syntactic and semantic analysis has been provided to the TTS manager.

From a general point of view, the scientific problem can be framed inside the need to develop models that are predictive for a speech synthesizer to be able to sound natural and expressive, getting as close as possible to human-like performance. This can only be achieved manipulating prosody so that the text read aloud sounds fully natural, informative and engaging or convincing. However, in order to achieve something closer to that, text understanding should be attained or some similar higher level semantic computation. As Xu (2011) puts it, "It will probably be a long time before anything close to that is developed, of course" (*ibi*, 94). Similar skeptical or totally negative opinions are expressed by Marc Huckvale (2002), when summarizing work he and his group have been carrying out for a long period over the project for an articulatory TTS called ProSynth. In fact, he bitterly criticizes current approaches at using machine learning and statistical processing to attain commercially viable results, as the one based on Unit Selection, because in this way no scientifically useful information will ensue towards the goal of increasing our knowledge of how human speech production works (*ibi*, 1261). The goal of speech synthesis, in this perspective would be that of "understanding how humans talk" rather than the one of replicating a human talker. In his opinion,

> We are not in any sense close to replicating how humans or HAL use speech because we are nowhere close to building a system with intentionality, with the need to communicate. The lack of expression or interest or emotion in the speech of current systems is due to the fact that the systems don't actually understand what they are saying, see no purpose to the communication, nor actually have any desires of their own. It makes no sense to add "expressiveness" or "emotion" to voice output from a mindless system (*ibi*, 1262).

This is what the majority of the systems are currently trying to do "Adding emotion to a mindless system", and they do that by focusing on specific words or expressions. Research on emotion from a perceptive and productive way has started long time ago (see Cahn, 1990), but exploded lately, say from 2000 onward. The bibliography is extensive and regards the six basic emotions (Disgust, Happiness, Sadness, Anger, Fear, Surprise, Neutral) as well as richer sets including Boredom, Despair, Anxiety, Contempt, Shame; Indignant, Pleasant. Emotion Generation is mainly achieved by statistical methods using different approaches: Formant and Articulatory Synthesis; Parametric Synthesis (HMMs); Diphone Synthesis; Unit Selection Synthesis.

Synthesis of expressive sounds is easy and can be extended easily to whole phrases by unit selection methods. This can be done by collecting separate databases for different emotions as suggested by Lakshmi et al.. This latter method seems the most promising in that it allows to address units of various size, starting from Phones, Halfphones but also Words, Phrases and entire Utterances, depending on the domain. Drawbacks are the need to improve on prosodic modeling and also to account for fine-grained differences in emotional content as clearly indicated by Nick Campbell in the conclusion of his talk for SSW6 (Speech Synthesis Workshop 6).

More linguistically based work on emotions has been documented by the group working at Loquendo (now acquired by Nuance). They report their approach based on the selection of Expression which is related to a small inventory of what they call "speech acts" – rather more Conversational and Argumentative or Dialogue acts. They implemented the acoustic counterpart of a limited, but rich, set of dialogue acts, including: refuse, approval/disapproval, recall in proximity, announce, request of information, request of confirmation, request of action/behaviour, prohibition, contrast, disbelief, surprise/ astonishment, regret, thanks, greetings, apologies, and compliments. In total, they managed to label and model accordingly some 500 different (expressive) utterances that can be used domain and context independently.

Work related to what we are trying to do is to be found in the field of storytelling and in experiments by the group from Columbia University working at MAGIC a system for the generation of medical reports. Montaño et al. present an analysis of storytelling discourse modes and narrative situations, highlighting the great variability of speech modes characterized by changes in rhythm, pause lengths, variation of pitch and intensity and adding emotion to the voice in specific situations. In their experiment on storytelling TTS they consider it as a first step towards modeling pitch, intensity and tempo. They found the following important possible triggers for prosody adjustment: a wide variety of speech variability; continual rhythm changes; including pauses of different duration; sometimes adding emotion to the voice. These have had as a consequence, variation of pitch and intensity, and the need to treat dialogue structures correctly. In other words, in order to make TTS for storytelling more natural and expressive they had to address those triggers and model rhythm variability, pauses and voice quality.

However, the most interesting approach related to ours is the one by the group of researchers from Columbia University, where we can find Julia Hirschberg and Kathy McKeown. In the paper by S. Pan, K. McKeown & J. Hirschberg (1999) they highlight the main objectives of their current work, i.e.

> The production of natural, intelligible speech depends, in part, on the production of proper prosody: variations in pitch, tempo and rhythm. Prosody modelling depends on associating variations of prosodic features with changes in structure, meaning, intent and context of the language spoken. This process involves selecting information that has potential to influence prosody, identifying correlations between this information and prosodic parameters through data exploration, and using learning algorithms to build prosody models from these data (*ibi*, 1420).

In fact, their attempt at using machine learning for building prosody models has been only partially achieved as we will see in detail below. Now the important part of their work is the one related to the concept-to-speech manager. As reported in their paper,

> The content planner uses a presentation strategy to determine and order content. It represents discourse structure, which is a hierarchical topic structure in MAGIC, discourse relations, which can be rhetorical relations, and discourse status, which represents whether a discourse entity is given, new or inferable and whether the entity is in contrast with another discourse entity. Most of the features produced at this stage have been shown to have influence on prosody: discourse structure can affect pitch range, pause and speaking rate (Grosz, Hirschberg, 1992); given/new/ inferable can affect pitch-accent placement (Hirschberg, 1993); a shift in discourse focus can affect pitch-accent assignment (Nakatani, 1998); and contrastive entities can bear a special pitch accent (Prevost, 1995) (*ibi*, 1422).

Further work towards predicting prosodic structure was published by Bachenko & Fitzpatrick, 1990, Delmonte & Dolci, 1991, and Wang & Hirschberg, 1992. The objective of their experiment was modeling ToBI prosody features, i.e. pitch accents, phrase accents, boundary tones and break indices. Given the fact that there are six pitch-accent classes, five break-index classes, three phrase-accent classes, and three boundary-tone classes, they come up with a total of 17 different features organized in four separate classes. The experiment was carried out on a corpus of spontaneous speech with some 500 dialogues on medical issues, which ended up by being reduced to 250 annotated dialogues.

In fact, the features they managed to annotate are just surface syntactic and semantic ones, and they are: (1) ID: the ID of a feature vector; (2) Lex: the word itself; (3) Concept: the semantic category of a content word; (4) SynFunc: the syntactic function of a word; (5) SemBoundary: the type of semantic constituent boundary after a word; (6) SemLength: the length, in number of words, of the semantic constituent associated with the current SemBoundary; (7) POS: the part-of-speech of a word; (8) IC: the semantic informativeness of a word, where in particular, the latter is – in our opinion – wrongly computed as a "semantic feature", being constituted by the logarithm of the relative frequency of a term in the corpus. The most disappointing fact was that they attempted to carry out a complete annotation but didn't succeed. In the paper they report their annotation efforts on the spontaneous-speech corpus which was automatically annotated with POS information, syntactic constituent boundaries, syntactic functions, and lexical repetitions, using approximations provided by POS taggers and parsers. It was also manually labelled with given/new/inferable information. But when it comes to semantic and discourse level information:

> We are still working on manually labelling discourse structure, discourse relations, and semantic abnormality... We are currently annotating the speech corpus with features closely related to meaning and discourse (*ibi*, 1426).

No further publication reports experiments with the complete annotation. And this is clearly due to difficulties inherent in the task. Now, this is what our system allows us to do, i.e. using discourse structure and relations to instruct the prosody manager to introduce the appropriate variation of prosodic parameters. According to ToBI features, this implies the ability to achieve:
– Juncture placing prediction;
– Phrase boundary tone prediction;
– Prominence prediction;
– Intonational Contour movement prediction.

To be more specific, given an input text the "Ideal System" will read it aloud using naturally sounding prosody, where:
– Phrasing is fully semantically consistent;
– Intonation varies according to structural properties of clauses in discourse and speaker intention;

- Prominence is assigned on the basis of novelty of topics and related events;
- Expressivity is intended to convey variations of attitude and mood as they are derived from deep subjective and affective analysis.

Our reformulation of ToBI features from general/generic into concrete and implemented analogical parameters for natural and expressive TTS will be achieved at the end of the paper. The correspondence between prosodic features and linguistic representation is the issue to cope with and will be presented here. Levels of intervention of syntactic-semantic and pragmatic knowledge include:

- syntactic heads which are quantified expressions;
- syntactic heads which are preverbal SUBJects;
- syntactic constituents that starts and ends an interrogative or an exclamative sentence;
- distinguish realis from irrealis mood;
- distinguish deontic modality including imperative, hortative, optative, deliberative, jussive, precative, prohibitive, propositive, volitive, desiderative, imprecative, directive and necessitative etc.;
- distinguish epistemic modality including assumptive, deductive, dubitative, alethic, inferential, speculative etc.;
- any sentence or phrase which is recognized as a formulaic or frozen expression with specific pragmatic content;
- subordinate clauses with inverted linear order; distinguishing causal from hypotheticals and purpose complex sentences;
- distinguishing parentheticals from appositives and unrestricted relatives;
- Discourse Structure to tell satellite and dependent clauses from main;
- Discourse Structure to check for Discourse Moves - Up, Down and Parallel;
- Discourse Relations to tell Foreground Relations from Backgrounds;
- Topic structure to tell the introduction of a new Topic or simply a Change at relational level.

There is general consensus on the need to provide "Natural Language Understanding" abilities to generate expressive TTS. We dub NLU as the ability to represent the semantics and pragmatics of an utterance, a text or a dialogue. What kind of semantics and pragmatics is then needed to produce expressive TTS? Do there exist state of the art systems that can generate such a representation?

We produced a number of computational models for a linguistically driven TTS in our research work for Digital Equipment in the '80s aiming at the Italian version of DecTalk (Delmonte et al., 1984; 1986). The scheme of interaction between linguistic processing and parameterization for TTS was fairly general at the time and included phonetics, phonology and syntax. No semantics was considered for lack of computer power and memory storage (Delmonte, 1982; 1985).

In our prosodic computational model (1986, figg. 1-4), all interactions between the two levels – the phonological and the grammatical one – are already clearly indicated. However, no semantic information would be made available except for syntactic structure computed with the help of a subcategorization lexicon of Italian and an ATN parser (1986, fig. 6).

Our current system, on the contrary, develops a fully specified semantic representation which is then responsible for the great majority of focus displacements and intonational movements. As is clarified above, semantics is an essential component and contributes most of the information needed for a TTS to be perceived as "expressive".

## 2. *Semantic Representation for TTS*

Systems that can produce an appropriate semantic representation for a TTS are not many at an international level but they can be traced from the results of a Shared Task organized by members of SigSem and are listed here below in the corresponding webpage http://www.sigsem.org/w/index.php?title=STEP_2008_ shared_ task:_comparing_semantic_representations. State of the art semantic systems are based on different theories and representations, but the final aim of the workshop was reaching a consensus on what constituted a reasonably complete semantic representation, as can be gathered from the questionnaire that has been used to grade each system.

Figure 1 - *System Architecture Modules for SPARSAR*

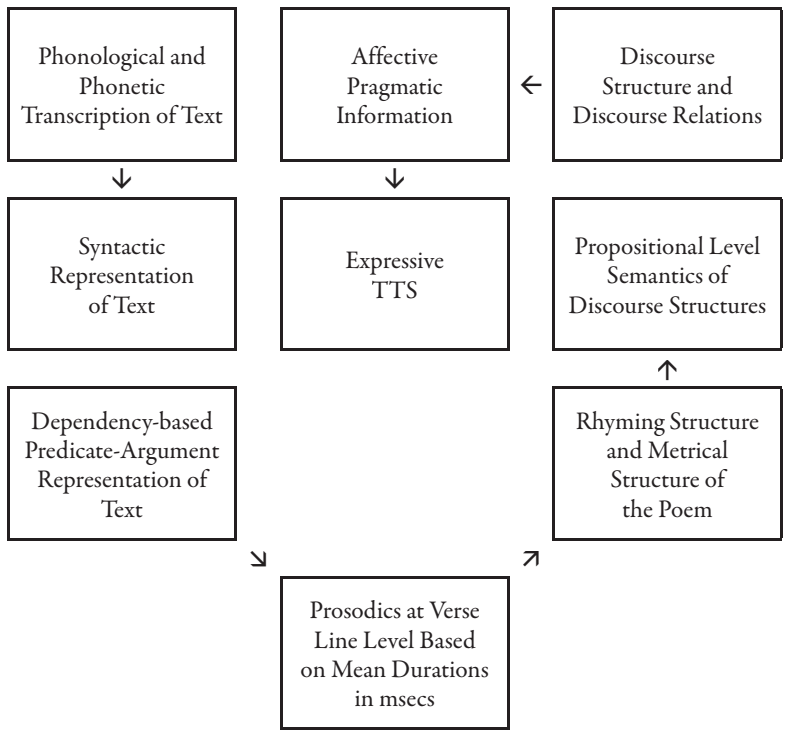

Semantics in our case not only refers to predicate-argument structure, negation scope, quantified structures, anaphora resolution and other similar items, it refers essentially to a propositional level analysis, then to a text and discourse level analysis. Propositional level semantic representation is the basis for discourse structure and discourse semantics contained in discourse relations. It also paves the way for a deep sentiment or affective analysis of every utterance, which alone can take into account the various contributions that may come from syntactic structures like NPs and APs where affectively marked words may be contained. Their contribution needs to be computed in a strictly compositional manner with respect to the meaning associated to the main verb, where negation may be lexically expressed or simply lexically incorporated in the verb meaning itself.

In Figure 1 we show the architecture of our deep system for semantic and pragmatic processing, in which phonetics, prosodics and NLP are deeply interwoven.

The system does the usual low level analyses before semantic modules are activated, that is tokenization, sentence splitting, multiword creation from a large lexical database. Then chunking and syntactic constituency parsing which is done using a rule-based recursive transition network: the parser works in a cascaded recursive way to include always higher syntactic structures up to sentence and complex sentence level. These structures are then passed to the first semantic mapping algorithm that looks for subcategorization frames in the lexica made available for English, including VerbNet, FrameNet, WordNet and a proprietor lexicon with most frequent verbs, adjectives and nouns, containing also a detailed classification of all grammatical or function words. This mapping is done following LFG principles, where c-structure is turned into f-structure obeying uniqueness, completeness and coherence. The output of this mapping is a rich dependency structure, which contains information related also to implicit arguments, i.e. subjects of infinitivals, participials and gerundives. It also has a semantic role associated to each grammatical function, that is used to identify the syntactic head lemma uniquely in the sentence. Finally it takes care of long distance dependencies for relative and interrogative clauses.

Now that fully coherent and complete predicate argument structures have been built, pronominal binding and anaphora resolution algorithms can be fired. Also coreferential processed are activated in the semantic level: they include a centering algorithm for topic instantiation and memorization that we do using a three-place stack containing a Main Topic, a Secondary Topic and an Potential Topic. In order to become a Main Topic, a Potential Topic must be reiterated and become persistent in the text.

Discourse Level computation is done at propositional level by building a vector of features associated to the main verb of each clause. They include information about tense, aspect, negation, adverbial modifiers, modality. These features are then filtered through a set of rules which have the task to classify a proposition as either objective/subjective, factual/nonfactual, foreground/ background. In addition, every lexical predicate is evaluated with respect to a class of discourse relations. Eventually, discourse structure is built, according to usual criteria of clause dependency where a clause can be classified either as coordinate or subordinate.

We have a set of four different moves to associate to each clause: root, down, level, up. We report here below semantic and discourse structures related to the poem by Sylvia Plath "Edge" which you can find here, http://www.poetryfoundation.org/poem/178970, and in the *Appendix*.

Figure 2 - *Propositional semantics for Edge*

| Clause No. | Funct/ Role | View | Factivity | Change | Relevance | Aspect | Pred | Tense | Disc_ Rel_ | Subject_ |
|---|---|---|---|---|---|---|---|---|---|---|
| 33 | coord/ prop | external | factive | null | background | activity | crackle | pres | narration | objective |
| 32 | coord/ prop | external | factive | null | background | activity | drag | pres | narration | objective |
| 26 | main/ prop | external | factive | culmintd | foregrnd | activity | use | perf | cause | objective |
| 21 | main/ prop | external | factive | null | background | activity | be | pres | narration | objective |
| 20 | main/ prop | external | factive | null | background | activity | have | pres | narration | objective |
| 16 | coord/ prop | external | factive | culmintd | foregrnd | activity | bleed | past | narration | objective |
| 15 | main/ prop | external | factive | null | background | activity | close | pres | narration | objective |
| 14 | main/ prop | external | factive | null | background | activity | fold | perf | narration | objective |
| 11 | main/ prop | external | factive | null | background | activity | bare | pres | narration | objective |
| 10 | main/ prop | external | factive | null | background | activity | be | pres | narration | objective |
| 9 | main/ prop | external | factive | null | background | activity | flow | pres | result | objective |
| 8 | main/ prop | external | factive | null | background | activity | have | pres | narration | objective |
| 7 | main/ prop | external | factive | null | background | activity | have | pres | narration | objective |
| 6 | main/ prop | external | factive | culmintd | foregrnd | activity | perfect | perf | result | objective |
| 5 | main/ prop | external | factive | culmintd | foregrnd | activity | say | past | narration | objective |
| 4 | main/ prop | external | factive | null | background | activity | seem | pres | narration | objective |
| 3 | main/ prop | external | factive | null | background | activity | wear | pres | narration | objective |
| 1 | xcomp/ prop | internal | factive | null | background | state | edge | nil | setting | objective |

In Figure 2, clauses governed by a copulative verb like BE report the content of the predication to the subject. The feature CHANGE can either be set to NULL, GRADED or CULMINATED: in this case Graded is not used seen that there no progressive or overlapping events.

Figure 3 - *Discourse level Semantics for Topic Hierarchy*

| Topic Type | Clause No. | Pred | Semant_ Id_ | M-Feats Per, Gen, Num | Semantic Inherent Feats | Semantic Role |
|---|---|---|---|---|---|---|
| main | 1 | edge | id1 | [3, neu, sing, | [abstrct, legal, nquant, objct], | theme_bound] |
| poten | 3 | illusion | id2 | [3, nil, nil, | [abstrct, inform, danger], | theme_bound] |
| poten | 3 | scroll | id3 | [3, mas, sing, | [abstrct, tecno], | goal] |
| poten | 3 | foot | id4 | [3, nil, nil, | [animat, body_part, objct], | theme_bound] |
| poten | 3 | smile | id5 | [3, mas, sing, | [activ, inform], | goal] |
| poten | 3 | toga | id6 | [3, nil, nil, | [body_part, objct], | theme_bound] |
| poten | 3 | dead_body | id7 | [3, mas, sing, | [objct, hum], | goal] |
| poten | 3 | necessity | id8 | [3, nil, nil, | [place, inform, state], | theme_bound] |
| poten | 3 | accom-plishment | id10 | [3, mas, sing, | [abstrct, chang, state], | goal] |
| main | 3 | woman | id2 | [3, fem, sing, | [any, relat, social, hum], | theme] |
| second | 15 | garden | id11 | [3, neu, plur, | [instit, objct, instrum], | agent] |
| poten | 15 | child | id12 | [3, neu, sing, | [any, activ, body_part, objct, relat, social, instrum, hum], | actor] |
| poten | 15 | serpent | id13 | [3, neu, sing, | [animt, objct, instrum], | theme] |
| poten | 15 | throat | id14 | [3, neu, plur, | [body_part, objct, instrum, hum], | loc_origin] |
| poten | 15 | stiffen | id16 | [3, neu, plur, | [instit], | goal] |
| poten | 15 | body | id7 | [3, neu, sing, | [abstrct, activ, body_part, inform, instit, place, objct, instrum, hum], | loc_direct] |
| poten | 15 | pitcher | id15 | [3, mas, sing, | [activ, inform, nquant, objct, relat, social, instrum, hum], | specif] |
| poten | 15 | milk | id17 | [3, neu, sing, | [body_part, edible, objct, hum], | specif] |
| poten | 15 | petal | id18 | [3, neu, plur, | [plant], | agent] |
| poten | 15 | flower | id19 | [3, neu, sing, | [plant, time], | theme] |
| poten | 15 | night | id20 | [3, neu, sing, | [state, time], | specif] |
| main | 21 | hood | id21 | [3, mas, sing, | [objct, instrum, hum], | loc_origin] |
| poten | 21 | moon | id22 | [3, neu, sing, | [event, place, objct, time], | experiencer] |
| poten | 29 | sort_of | id23 | [3, nil, nil, | [abstrct, activ, inform, relat, social, state, tecno, hum], | attr] |

In Figure 3. below, we see topics of discourse as they have been computed by the coreference algorithm, using semantic indices characterized by identifiers starting with ID. Every topic is associated to a label coming from the centering algorithm: in particular, WOMAN which is assigned ID id2 reappears as MAIN topic in clauses marked by n. 15. Also BODY reappears with id7. Every topic is associated to morphological features, semantic inherent features and a semantic role. Eventually, the final computation which concerns Discourse Structure is this one:

Figure 4 - *Discourse Semantics for Discourse Structures*

| Sent_ No. | Clause No. | Subject_ | Disc_ Rel_ | Tense | Pred | Relevance | Disc_ Move | Disc_Struct_ Attach_Node |
|-----------|-----------|----------|------------|-------|------|-----------|------------|--------------------------|
| edge_7 | 39 | objective | narration | pres | crackle | background | level | down(31-39)) |
| edge_7 | 38 | objective | narration | pres | drag | background | level | down(31-39)) |
| edge_6 | 31 | objective | cause | perf | use | foreground | up | to(1-31)) |
| edge_5 | 25 | objective | narration | pres | moon | background | level | down(18-25)) |
| edge_5 | 24 | objective | narration | pres | have | background | level | down(18-24)) |
| edge_5 | 23 | objective | narration | pres | stare | foreground | down | down(18-23)) |
| edge_4 | 18 | objective | narration | past | bleed | foreground | up | to(1-18)) |
| edge_4 | 17 | objective | narration | perf | fold | background | level | level(11-17)) |
| edge_4 | 15 | objective | circumstance | pres | stiffen | background | level | level(11-16)) |
| edge_3 | 11 | objective | narration | pres | it | background | level | level(7-11)) |
| edge_3 | 10 | objective | narration | pres | come | foreground | level | level(7-10)) |
| edge_2 | 7 | objective | result | pres | flow | background | down | down(1-7)) |
| edge_2 | 5 | objective | narration | past | say | foreground | up | to(1-5)) |
| edge_2 | 4 | objective | narration | pres | seem | background | level | level(1-4)) |
| edge_2 | 3 | objective | narration | pres | wear | background | level | level(1-3)) |
| edge_2 | 6 | objective | narration | perf | perfect | foreground | down | down(1-6)) |
| edge_1 | 1 | objective | setting | nil | edge | background | up | down(nil-1)) |

Movements in the intonational contours are predicted to take place when FOREGROUND and UP moves are present in the features associated to each clause. At discourse level computation, Background and Foreground are associated to opposing semantic analyses. In the case of Foreground, a new event is taking place or a new character appears on the scene. Thus Foreground is based on temporal and aspectual analysis, anaphoric and coreference analysis. These in turn are based on propositional level analysis as explained below. In addition, discourse structures and relations are responsible for a choice between Back and Foreground, that is the meaning of the governing predicates of each clause.

## 3. *From Poetry to Story Reading*

We referred to a poem in the previous section because in fact we will be using rules associated to poetry prosodic mapping in our work on story reading. We assume that reading a story aloud requires the reader to organize pauses in such a way that expressivity and meaning is preserved. This process is usually referred to as Breath Group organization: a breath group is a well-formed group of words conveying a concept or a meaning, and we would like to compare it to a line in a poem. Poems are organized into lines and stanzas, while stories usually have punctuation to mark main concepts and introduce pauses. Punctuation however is not sufficient in itself and does not always guarantee meaning coherence. We already noted that Commas are highly ambiguous and may be used for a whole set of different functions in discourse. So eventually what we can actually trust are Breath Groups. Continuing our comparison with poems, lines may be end-stopped or enjambed when they run on the following line or stanza. The same may happen with Breath Groups, they may be end-stopped or enjambed and require a different prosodic setup.

We will then define Breath Groups as syntactically and semantically coherent units coinciding with an Intonation Phrase in ToBI terms: IPs are characterized by different tones, possible boundary tones and break indices. Pitch Accents are associated to word accents which are present in our phonetic representation: except that only syntactic heads are associated with Pitch Accents, dependents are demoted.

### 3.1 Implementing the Rules for Expressive TTS

Let's now look at one example, a short story by Aesop, "Bellying the Cat" that can be found here, http://www.taleswithmorals.com/aesop-fable-belling-the-cat.htm. At first we show the decomposition of the story into Breath Groups and then the mapping done by the Prosodic Manager (hence PM):

Table 1 - *Decomposition of the text into Breath Groups*

| |
|---|
| long_ago ß |
| the mice had a general council ß |
| to consider what measures they could take to outwit their common enemy ß |
| the cat ß |
| some said this ß |
| and some said that ß |
| but at_last a young mouse got_up ß |
| and said he had a proposal ß |
| to make ß |
| which he thought would meet the case ß |
| ß |
| you will all agree ß |
| said he ß |
| that our chief danger consists in the sly ß |
| and treacherous manner ß |
| in which the enemy approaches us ß |

```
                              ß
                           now ß
         if we could receive some signal of her approach ß
               we could easily escape from her ß
                         i venture ß
                        therefore ß
           to propose that a small bell be procured ß
     and attached by a ribbon round the neck of the cat ß
                     by_this_means ß
          we should always know when she was about ß
                and could easily retire ß
            while she was in the neighborhood ß
                              ß
          this proposal met with general applause ß
              until an old mouse got_up ß
                       and said ß
                              ß
               that is all very_well ß
              but who is to bell the cat ß
                              ß
          the mice looked at one_another ß
                 and nobody spoke ß
              then the old mouse said ß
                              ß
                     it is easy ß
          to propose impossible remedies ß
```

A first set of the rules to map the story into this structures are reported below. The rules are organized into two separate sets: low level and high level rules. Here are low level ones:

### 3.1.1 Breath Group Creation Rules

– Follow punctuation first, but check constituent length;
– Rule for Coordinate Structures;
– Rule for Subordinate Clauses;
– Rule for Infinitival Complements;
– Rule for Complement Clauses;
– Rule for Relative Clauses;
– Rule for Between Subject and VerbPhrase;
– Rule for AdverbialPhrase but only beginning of Clause;
– Rule for Obligatory complements followed by adjuncts – with long constituents.

The high level corresponds to the recursive level. Recursive rules are associated with complex sentences and with Coordinate, Subordinate and Complement clauses. These rules need to be conceived in their recursive structure when implemented in a real algorithm, in particular, they will reorganized as follows.

Here below is the mapping of the text above into analogical phonetic acoustic correlates of pitch, speaking rate and intensity, and pauses. Here PBAS is used for F0 rise/fall instructions; RATE for speaking rate; VOLM for energy or intensity; SLNC stands for silence, hence pauses and RSET for reset all parameter to the default value. However, since we intend to refer to ToBI framework for prosody annotation, we will then match every new feature we inserted in the text with the inventory made available by the original ToBI manual – see below. This analogical version can be copy/pasted into a TextEdit file and spoken aloud by Apple TTS.

[[pbas 38.000; rate 160; volm +0.5]]Bellying the cat, a story by Aesop . [[slnc 400]],[[rset 0]] [[pbas 44.000; rate 140; volm +0.3]][[pbas 36.000; rate 110; volm -0.2]]Long ago [[rset 0]] , the mice had a general council to [[pbas 36.000; rate 110; volm +0.5]]consider[[slnc 50]],[[rset 0]] what measures they could [[pbas 36.000; rate 110; volm +0.5]]take[[slnc 30]],[[rset 0]] to [[slnc 100]][[pbas 40.000; rate 150; volm +0.5]]outwit their common [[pbas 38.000; rate 130; volm +0.3]]enemy[[slnc 200]],[[rset 0]] , the [[pbas 38.000; rate 130; volm +0.3]]cat[[slnc 200]],[[rset 0]] .

Some [[pbas 36.000; rate 110; volm +0.5]]said[[slnc 50]],[[rset 0]] [[pbas 38.000; rate 130; volm +0.3]]this[[slnc 200]],[[rset 0]] , [[slnc 100]]and some said [[pbas 38.000; rate 130; volm +0.3]]that[[slnc 200]],[[rset 0]] ;

[[pbas 36.000; rate 110; volm +0.5]]but[[slnc 30]],[[rset 0]] at_last a young [[rate 130; volm +0.5]]mouse got_up [[slnc 100]]and said he had a proposal to [[pbas 38.000; rate 130; volm +0.3]]make[[slnc 200]],[[rset 0]] , which he thought [[slnc 100]][[pbas 50.000; rate 120; volm +0.5]]would meet the [[pbas 38.000; rate 130; volm +0.3]]case[[slnc 200]],[[rset 0]] .

" You will [[rate 130; volm +0.5]]all agree " , said [[pbas 38.000; rate 130; volm +0.3]]he[[slnc 200]],[[rset 0]] , " that our chief danger consists in the [[pbas 36.000; rate 110; volm -0.2]]sly and treacherous [[rset 0]] manner in which the enemy approaches [[pbas 38.000; rate 130; volm +0.3]]us[[slnc 200]],[[rset 0]] " .

Now , if we [[slnc 100]][[pbas 50.000; rate 120; volm +0.5]]could receive some signal of [[rate 130; volm +0.5]]her approach , we could easily escape from [[pbas 38.000; rate 130; volm +0.3]]her[[slnc 200]],[[rset 0]] .

I [[pbas 38.000; rate 130; volm +0.3]]venture[[slnc 200]],[[rset 0]] , [[pbas 38.000; rate 130; volm +0.3]]therefore[[slnc 200]],[[rset 0]] , to [[pbas 36.000; rate 110; volm +0.5]]propose[[slnc 50]],[[rset 0]] that a small bell [[slnc 100]][[pbas 50.000; rate 120; volm +0.5]]be procured , [[slnc 100]]and attached by a ribbon round the neck of the [[pbas 38.000; rate 130; volm +0.3]]cat[[slnc 200]],[[rset 0]] .

by this means we should always know [[slnc 100;pbas 48.000; rate 150; volm +0.3]] when she was [[pbas 38.000; rate 130; volm +0.3]]about[[slnc 200]],[[rset 0]] , [[slnc 100]]and could easily [[pbas 36.000; rate 110; volm +0.5]]retire[[slnc 50]],[[rset 0]] while she was in the [[pbas 38.000; rate 130; volm +0.3]]neighborhood[[slnc 200]],[[rset 0]] " .

This proposal [[slnc 100]][[pbas 50.000; rate 120; volm +0.5]]met with general [[pbas 38.000; rate 130; volm +0.3]]applause[[slnc 200]],[[rset 0]] , until an old mouse [[slnc 100]][[pbas 50.000; rate 120; volm +0.5]]got up [[slnc 100]]and [[pbas 38.000; rate 130; volm +0.3]]said[[slnc 200]],[[rset 0]] : [[pbas 48.000; rate 130; volm +0.9]]

" That is [[rate 130; volm +0.5]]all very_well , but [[pbas 54.000; rate 170; volm +0.3]]who is to bell the cat[[slnc 300; pbas 54.000; rate 170; volm +0.3]] ? [[rset 0]]

" the mice looked at one another [[slnc 100]]and [[rate 110; volm +0.3]]nobody[[slnc 100]],[[rset 0]] [[pbas 38.000; rate 130; volm +0.3]]spoke[[slnc 200]],[[rset 0]] .

[[pbas 54.000; rate 170; volm +0.3]]Then the old mouse said : [[pbas 48.000; rate 130; volm +0.9]]
" it is [[pbas 40.000; rate 140; volm +0.3]]easy[[slnc 30]],[[rset 0]] to propose [[pbas 36.000; rate 110; volm -0.2]]impossible remedies [[rset 0]] " .

## 3.2 ToBI features Implemented

We will now discuss the use of Pierrehumbert's inventory of Tones and Break Indices, in relation to its actual application in real texts reading. We tried to map the analogical features required by the prosodic manager into ToBI features but we soon discovered that more features were needed. This is different from trying to predict ToBI indices on a statistical basis, as can be found in extended number of works in the literature – we refer here to one such paper, by Lee, Kim & Lee, 2001. We shall start from Break Indices which amount to 5, starting from 0 to 4 included. Here below the list of features from AME (American) ToBI inventory and their interpretation according to the original manual which can be found at http://www.speech.cs.cmu.edu/tobi/:

0: word boundary apparently erased;
1: typical between-word disjuncture within a phrase;
2: mismatched inter-word disjuncture within a phrase;
3: end of an intermediate phrase;
4: end of an intonational phrase.

We assume that in fact BI 0 is a very special and peculiar index and covers an aspect of prosody which has no general practical application. As for BI 2 we will use it to cover one of the phenomena indicated in the manual, that the idea to indicate a stronger sense of disjuncture than 1, for careful deliberation as in the comment reported in the footnote. From what we can see, we come up with two types of BIs: 3 and 4 are also intonationally related and regard phrase and sentence level prosody. BI 1 and B 2 are to be regarded as pertaining to word level and to possible internal constituent or internal phrase disjuncture. The latter BIs have no effect on the intonational contour. In terms of our analogical parameterization, the former two indices require a **_reset_** at the end that accompany the **_silence_**, the latter two have no **_reset_**. However, when we completed our experimentation, we discovered that many more disjunctures needed to be marked and for different purposes than the ones indicated in the original ToBI manual. Here is the list we have drawn at the end of our annotation task.

| | |
|---|---|
| [[slnc 300]],[[rset 0]] | **BI-4** |
| [[slnc 200]],[[rset 0]] | **BI-3** |
| [[slnc 100]] | **BI-2** |
| [[slnc 30]],[[rset 0]] | **BI-32** |
| [[slnc 50]],[[rset 0]] | **BI-33** |
| [[slnc 100]],[[rset 0]] | **BI-23** |
| [[slnc 300]] | **BI-22** |
| [[slnc 400]] | **BI-44** |
| [[rate 110; volm +0.3]] | **<slow down** |
| [[rate 130; volm +0.5]] | **<slow down** |

There are different 3 breaks, the reason for that is due to the use of the break in presence of end of Breath Group, with punctuation (BI 3) and without punctuation. The latter case is then split into two subcases, one in which the last word – a syntactic head – is followed or not by a dependent word, hence 33 and 32 respectively are the indices we used. We also use 44 for the separation of the title from the text. Finally 23 is a break with a reset between constituents of a specific type, quantifiers. Then we have two slow down commands, where again one precedes quantifiers, and the other for all syntactic heads, indicating end of Breath Groups. Coming now to tones and accents, the list from the manual includes 14 different combinations. Just to recollect the meaning of some specific label like *downstep*, we report their comment: "downstepped H star: a downstepped H pitch accent indicates that the tone of the prominent syllable is realized by a perceptually lower f0 than that of an immediately preceding High tone: the tone has 'stepped down' from the preceding High."

It follows that since this is High tone realized with a Lowering movement there is no downstepped L star index. Also consider Bitonal pitch accents which are two tone as part of the same pitch accent where the use of + indicates that the two tones are associated and form a single unit: a complex bitonal accent. We assume that a bitonal pitch accents can only be used for special cases of surprise or for expressing contrast. So they would be rather rare in real texts and dialogues. We then decided to port it to more common structures and will show one such usage below which however does not appear in the original list of the manual.

In our new ToBI list, we find downstepped tones to be rather common. They would be necessary every time a sequence of two or more clauses are expressed in a sequence and are interconnected by some discourse relation, usually through the use of a discourse marker.

We report here below a table which includes the new classification where we associate Analogical Parameters used for TTS with their symbolic representation in ToBI.

Table 2 - *Description of Tones Pitch Accents with their corresponding Analogical Parameterization*

| Description | Analogical Pramaterization | ToBI |
| --- | --- | --- |
| Beginning of Text for the title | pbas 38.000; rate 160; volm +0.5 | **H*-L** |
| End of Breath Group sentence internally | pbas 38.000; rate 130; volm +0.3 slnc 200, rset 0 | **H*-L%** **BI 3** |
| Beginning of sentence when expressing an up movement and a foreground relevance in discourse structure | pbas 44.000; rate 140; volm +0.3 | **H*-H** |
| Beginning of sentence when expressing an up movement and a foreground relevance in discourse structure and it is preceded by a paragraph boundary | pbas 54.000; rate 170; volm +0.3 | **H*-H 1** |
| Sentence internal and a Breath Group boundary | pbas 40.000; rate 140; volm +0.3 | **H*-L% 1** |

| Description | Analogical Pramaterization | ToBI |
|---|---|---|
| End of Breath Group and syntactic head | pbas 36.000; rate 110; volm +0.5<br>slnc 50, rset 0 | **L-L%**<br>**BI 33** |
| Sentence internal and a foreground relevance in discourse structure | pbas 40.000; rate 150; volm +0.5 | **H*-L** |
| Sentence internal and an adjunct clause with foreground relevance in discourse structure | pbas 50.000; rate 120; volm +0.5 | **H-H* 2** |
| Sentence internal and an adjunct clause with background relevance in discourse structure | pbas 40.000; rate 120; volm +0.5 | **H-H* 4** |
|  | pbas 38.000; rate 130; volm +0.3<br>slnc 200, rset 0 | **H*-L% 2**<br>**BI 3** |
| Direct speech Breath Group boundary + Exclamative | pbas 54.000; rate 170; volm +0.3 | **BI 44** |
|  |  | **H*-H%** |
| SAD affective tone associated to a phrase or word | pbas 36.000; rate 110; volm -0.2 | **L*-L%** |
|  | rset 0 |  |
| Sentence internal and a discourse marker indicates beginning of a subordinate clause | pbas 48.000; rate 150; volm +0.3 | **H*-H 3** |
|  | pbas 44.000; rate 140; volm +0.3] | **H-!H* 2** |
| Sentence internal and a coordinate clause with foreground relevance in discourse structure | pbas 50.000; rate 120; volm +0.5 | **H*-H 2** |
|  | pbas 44.000; rate 140; volm +0.3 | **H-!H* 2** |
| Direct Speech + Elaboration or Explanation | pbas 54.000; rate 170; volm +0.3 | **H*-H 1** |
|  | pbas 50.000; rate 160; volm +0.5 | **H-!H* 1** |
| Sentence internal and a discourse marker indicates beginning of a subordinate clause | pbas 48.000; rate 150; volm +0.3 | **H-!H*** |
|  | pbas 44.000; rate 140; volm +0.3] |  |
| Sentence internal and a coordinate clause with foreground relevance in discourse structure | pbas 50.000; rate 120; volm +0.5 | **H-!H*** |
|  | pbas 44.000; rate 140; volm +0.3 |  |
| Declarative sentence with a Resultative Infinitival | slnc 100; pbas 40.000; rate 150; volm +0.5 | **H-!L*** |
|  | slnc 100; pbas 38.000; rate 150; volm +0.5 |  |
| Split Exclamative | pbas 54.000; rate 170; volm +0.3<br>pbas 36.000; rate 110; volm -0.2 | **H*+L%** |
|  | rset 0 |  |
| Exhortative | pbas 57.000; rate 170; volm +0.5 |  |
|  | pbas 36.000; rate 170; volm +0.5 | **H*+L-** |

| Description | Analogical Pramaterization | ToBI |
|---|---|---|
| | pbas 24.000; rate 130; volm +0.5 | |
| | pbas 60.000; rate 150; volm +0.5 | **!L+H*%** |
| | slnc 100, rset 0 | **BI 23** |

Here below we explain some of the linguistic pragmatic description we associated to a combination of parameters like Exhortative. This is associated with the complex combination of tones used to express the specific meaning associated to utterances like "Come on, baby", which is to be regarded as an affective invitation. Here below is the set of instructions starting from the command that looks for the expression in the input string, the first sequence of parameters constituted by a bitonal pitch accent made of a very high tone followed by a low one, H*+L- placed on the expression "Come on" and a following tail made by the word "baby" where on the contrary we have a very special case of !L+H*% where the Low tone is upstepped with respect to the previous low.

```
checkfrozen(W,W1,Sent),
['[[pbas 57.000; rate 170; volm +0.5]]',W,
'[[pbas 36.000; rate 170; volm +0.5]]','',W1,'']
H*+L-

Sent=[Dear,'|Rest],
checkdear(Dear),
'[[pbas 24.000; rate 130; volm +0.5]]',Dear,
!L+H*%
'[[pbas 60.000; rate 150; volm +0.5]]',
'[[slnc 100]],[[rset 0]]']
```

We think in this way to have increased the inventory organized theoretically by the ToBI group, which has shown to be useful when it comes to reorganizing the information needed in a real implementation.

### 3.3 One specific case: downstepped Direct Speech

Consider now the case of another of the fables by Aesop we worked on – The Fox and the Crow, that can be found here, http://www.taleswithmorals.com/aesop-fable-the-fox-and-the-crow.htm. In this story the main character introduced by the narrator, starts his speech with an exclamative sentence and then continues with some explanation and elaborations. These discourse structures need to be connected to the previous level of intonation. This requires receiving information at clause level from the discourse level, in order to allow for the appropriate continuation. In particular, this is done by:

– detecting the presence of Direct Speech by both verifying the presence of a communication verb governor of a sentence started by the appropriate punctuation mark, inverted commas. This same marker will have to be detected at the end of direct speech. The end may coincide with current sentence or a number of additional sentences might be present as is the case at stake. The current reported speaker continues after the exclamative with a series of apparently neutral declarative sentences, which can be computed as explanations and elaborations. But they all depend from the exclamative and need to be treated accordingly at discourse level.

To work at discourse level, the system has a switch called "point of view" which takes into account whose point of view is reported in each sentence. The default value for a narrative text would be the "narrator" whenever the sentence is reported directly with no attribution of what is being said. When direct speech appears, the point of view is switched to the character whom the sentence has been attributed to. This switch is maintained until the appropriate punctuation mark appears. So eventually, it is sufficient for the Prosodic Manager (PM) to take the current point_of_view under control. If it is identical to the previous one, nothing happens. If it changes to a new holder and it is marked with direct speech, the algorithm will be transferred to a different recursive procedure which will continue until point_of_view remains identical. This new procedure allows the PM to assign downstepped intonational contours as shown here below. In this fragment, we also mark the presence of a word – HUE – which is wrongly pronounced by Apple synthesizer and requires activating the exceptional phonetic conversion.

> "What a noble bird I see **BI-3** above me **BI-22 H\*-H-1** ! **BI-2 H-!H\*-1** Her beauty is without **H\*-L%** equal **BI-3** , **H\*-L** the [[inpt PHON]]hUW[[inpt TEXT]] of her plumage **H\*-H-3** exquisite **BI-2 . H-!H\*-1** If only her voice is **BI-2** as sweet **BI-2** as her **BI-2 H-!H\*-1** looks are **H\*-L** fair **BI-3** , she **BI-2 H-H\*-2** ought **L\*-L%** without doubt [[rset 0]] to be Queen of the **H\*-L%-2** Birds **BI-3** ."

In case this information was not made available to the PM, the result would have been the following.

> "What a noble bird I see **BI-3** above me **BI-22 H\*-H-1** ! **BI-2 H-!H\*-1! H\*-L%** equal **BI-3** , **H\*-L** the [[inpt PHON]]hUW[[inpt TEXT]] of her plumage **H\*-H-3** exquisite **BI-2** .
> If only her voice is **BI-2** as sweet **BI-2** as her **BI-2 H-!H\*-1** looks are **H\*-L** fair **BI-3** , she **BI-2 H-H\*-2** ought **L\*-L%** without doubt [[rset 0]] to be Queen of the **H\*-L%-2** Birds **BI-3** ."

## 5. *Evaluation and Conclusion*

The system has undergone extensive auditory evaluation by expert linguists. It has also been presented at various demo sessions always receiving astounded favourable comments (see Delmonte, Bacalu, 2013; Delmonte, Prati, 2014; Delmonte, 2015). The evaluation has been organized in two phases, at first the story is read by Apple TTS directly from the text. Then the second reading has been done by the system and a comparison is asked of the subject listening to it. In the future we intend to produce an objective evaluation on a graded scale using naïve listeners English native speakers. We will be using the proposal in Xu (2011: 95), called MOS, or Mean Opinion Score, with a five-level scale: 5-Excellent, 4-Good, 3-Fair, 2-Poor, 1-Bad, with the associated opinions: 5-Imperceptible, 4-Perceptible but not annoying, 4-Slightly annoying, 2-Annoying, 1-Very annoying.

In this paper we presented a prototype of a complete system for expressive and natural reading which is fully based on internal representations produced by syntactic and semantic deep analysis. The level of computation that is mostly responsible for prosodic variations is the discourse level, where both discourse relations, discourse structures, topic and temporal interpretation allow the system to set up an interwoven concatenation of parameters at complex clause and sentence level. Pragmatically frozen phrases and utterances are also fully taken into account always at a parameterized level. Parameters have been related to ToBI standard set and a new inventory has been proposed. The system is currently working on top of Apple TTS but will be soon ported to other platforms. It is available for free download at a specific website: sparsar.wordpress.com.

## *Appendix:* Edge *by Sylvia Plath*

The woman is perfected.
Her dead

Body wears the smile of accomplishment,
The illusion of a Greek necessity

Flows in the scrolls of her toga,
Her bare

Feet seem to be saying:
We have come so far, it is over.

Each dead child coiled, a white serpent,
One at each little

Pitcher of milk, now empty.
She has folded

Them back into her body as petals
Of a rose close when the garden

Stiffens and odors bleed
From the sweet, deep throats of the night flower.

The moon has nothing to be sad about,
Staring from her hood of bone.

She is used to this sort of thing.
Her blacks crackle and drag.